\documentclass[12pt]{article}

\usepackage{setspace}
\usepackage{tablefootnote}
\usepackage[table]{xcolor}
\usepackage{csvsimple}
\usepackage{siunitx}
\usepackage{booktabs}
\usepackage{longtable}
\usepackage{lscape}
\usepackage[hidelinks]{hyperref}

\usepackage{graphicx} 
\usepackage{setspace}
\usepackage{fullpage}
\usepackage{amsmath}
\usepackage{amssymb}
\usepackage{amsthm}
\usepackage{booktabs}
\usepackage{enumitem}
\usepackage{placeins}
\usepackage{titlesec}
\usepackage{amsfonts}
\usepackage{mathrsfs}
\usepackage{appendix}

\titlespacing*{\subsection}{0pt}{0pt}{0pt}
\titlespacing*{\section}{0pt}{0pt}{0pt}
\titlespacing*{\subsubsection}{0pt}{0pt}{2pt}

\makeatletter
\def\thm@space@setup{\thm@preskip=0pt \thm@postskip=0pt}
\makeatother

\usepackage[authoryear]{natbib}
\usepackage{comment}
\usepackage{subfigure}
\usepackage{caption}
\usepackage{multicol}
\usepackage{multirow}
\usepackage{apptools}
\usepackage{subcaption}
\usepackage{booktabs}
\usepackage{comment}
\usepackage{xcolor}
\usepackage{tabularx}
\usepackage{adjustbox}

\usepackage{tikz}
\usepackage{bbm}
\usepackage{adjustbox}
\usepackage{makecell}
\usepackage{diagbox}
\usepackage{float}
\usepackage{graphicx}
\usepackage{caption}
\theoremstyle{plain}

\newtheorem{prop}{Proposition}

\newtheorem{lemma}{Lemma}

\DeclareMathOperator{\Cov}{Cov}
\DeclareMathOperator{\Var}{Var}

\newcommand{\E}{\mathbb{E}}
\title{An Accounting Identity for Algorithmic Fairness}
\date{\today}
\author{Hadi Elzayn \and Jacob Goldin}

\begin{document}

\maketitle

\begin{abstract}\noindent
We derive an accounting identity for predictive models that links accuracy with common fairness criteria. The identity shows that for globally calibrated models, the weighted sums of miscalibration within groups and error imbalance across groups is equal to a “total unfairness budget.” For binary outcomes, this budget is the model’s mean-squared error times the difference in group prevalence across outcome classes. The identity nests standard impossibility results as special cases, while also describing inherent tradeoffs when one or more fairness measures are not perfectly satisfied. The results suggest that accuracy and fairness are best viewed as complements in binary prediction tasks: increasing accuracy necessarily shrinks the total unfairness budget and vice-versa. Experiments on benchmark data confirm the theory and show that many fairness interventions largely substitute between fairness violations, and when they reduce accuracy they tend to expand the total unfairness budget. The results extend naturally to prediction tasks with non-binary outcomes, illustrating how additional outcome information can relax fairness incompatibilities and identifying conditions under which the binary-style impossibility does and does not extend to regression tasks.

\end{abstract}
\doublespacing

\section{Introduction}

As predictive algorithms are increasingly used to allocate benefits and burdens in modern society, a large and growing academic literature studies how these tools may inadvertently reinforce existing patterns of societal inequality. A central theme in this literature is that there are many plausible ways to formalize ``algorithmic fairness'' with respect to some protected group. Among the most commonly studied criteria are (i) \emph{calibration within groups}, (ii) \emph{balance across groups on the positive class}, and (iii) \emph{balance across groups on the negative class}. An influential set of results establishes general conditions under which these intuitively appealing criteria are mutually incompatible: except in degenerate cases, no predictive model can perfectly satisfy each fairness requirement in a given application \citep{chouldechova2017fair,kleinberg2016inherent}.

Faced with this impossibility, a natural question is how these aspects of fairness constrain one another in the feasible region---that is, when at least one fairness metric is violated. We study this question for globally calibrated predictive models. Our main point of departure from the prior literature is to study the \emph{magnitude} of departures from perfect fairness along each dimension, rather than treating fairness as an all-or-nothing property. This change in focus is motivated by normative and practical considerations. Normatively, the concerns that motivate common algorithmic fairness criteria are themselves continuous in nature: for example, if one worries that an algorithm systematically concentrates inflated risk scores on members of a disadvantaged group, that concern is generally more severe when the induced error rate imbalance is larger. In the language of moral philosophy, the fairness dimensions on which we focus---miscalibration by group and imbalance across groups---are better viewed as \emph{scalar properties} of an algorithm than as \emph{binary judgments} \citep{alexander2008scalar}. The practical motivation to focus on the degree of fairness violations is that because perfect satisfaction of all fairness criteria is typically infeasible, the relevant question for design and regulation is \emph{how much} unfairness of each type an algorithm induces.

We show that the fairness metrics on which we focus are linked by a simple accounting identity. The identity states that for a globally calibrated model, the weighted sum of unfairness---i.e., the degree to which each fairness criterion is violated (on net)---is equal to a quantity that we refer to as the \emph{total unfairness budget}. The weights in this identity are determined by where the population mass and cross-group overlap lie with respect to outcome levels and risk scores.

In the identity's most general formulation, the total unfairness budget is determined by the association between a model's prediction error and the variation in group prevalence driven by group differences in outcomes. Total unfairness against a group is larger when members of that group are disproportionately concentrated in outcome regions where the model's prediction errors tend to be large. The accounting identity shows that this association determines how much total unfairness must be allocated across different forms of unfairness in any globally calibrated model.

In the case of a binary outcome, the total unfairness budget takes a particularly simple form: the product of the difference in group prevalence across outcome classes and the model's mean-squared error (equivalently, the Brier score). The identity equates this total unfairness budget to the weighted sum of three familiar forms of unfairness---miscalibration by group, imbalance on the positive class, and imbalance on the negative class. Consequently, improving predictive accuracy mechanically shrinks the total unfairness budget that must be allocated across the various forms of unfairness.

The accounting identity we derive yields a number of useful insights for studying globally calibrated predictive models. Beginning with the case of a binary outcome:
\begin{itemize}[nosep]
    \item The accounting identity provides a simple and general relationship between common fairness metrics, baseline group differences, and model accuracy.
    \item The identity strengthens and generalizes the impossibility result of \citet{kleinberg2016inherent} by characterizing the feasible degree of fairness violations away from the boundary case in which each fairness condition is perfectly satisfied. 
    \item A direct implication is that fairness interventions that preserve model accuracy (for globally calibrated models) do not change the total unfairness budget: they can only substitute between different forms of unfairness. Conversely, fairness adjustments that reduce accuracy not only reallocate unfairness across dimensions, but also expand the total unfairness budget that must be allocated.
    \item The identity therefore clarifies the relationship between accuracy and fairness. While there can be tradeoffs between accuracy and \emph{particular} fairness dimensions, increasing accuracy necessarily reduces the total unfairness budget. In this sense, accuracy and fairness are complements rather than substitutes for globally calibrated binary prediction models.
\end{itemize}

The general version of our accounting identity also yields insights for settings in which the outcome is non-binary:
\begin{itemize}[nosep]
    \item We provide natural generalizations of the fairness metrics commonly applied in binary classification.
    \item The general accounting identity makes explicit how these fairness metrics are jointly determined by the distribution of group membership and the distribution of prediction errors across the outcome space.
    \item The identity highlights that the fairness incompatibility results for binary outcomes need not extend mechanically to regression settings. We provide a stylized counterexample in which both calibration within groups and balance across groups are satisfied by a non-oracle regressor even when mean outcomes differ by group. We then give a sufficient condition under which such counterexamples exist, as well as a sufficient condition under which the impossibility result does extend to regression.
\end{itemize}

\subsection*{Related Work}
The classic impossibility results relating to groupwise calibration and balance are due to \citet{chouldechova2017fair} (focusing on binary classifiers) and \citet{kleinberg2016inherent} (focusing on risk scores).\footnote{Between these polar focuses, \citet{reich2020possibility} show that groupwise calibration of a risk score can be compatible with equal error rates of a downstream classifier by enforcing the two criteria at different stages of the decision pipeline.} A large follow-on literature has clarified that these tensions reflect a structural incompatibility between “sufficiency”-type criteria (e.g., predictive parity / calibration within groups) and “separation”-type criteria (e.g., equalized odds / error rate balance across groups) when groups differ in baseline outcome prevalence. See  \citet{hardt2016equality}, as well as surveys by \citet{sep-algorithmic-fairness}, \citet{mitchell2021algorithmic} and \citet{mehrabi2021survey}. 

Closely related to our work is \citet{pleiss2017fairness}, who characterizes the trade-off between false-positive and false-negative errors under the constraint that perfect calibration within groups holds exactly. Analogously, our accounting identity implies how, when a risk score satisfies groupwise calibration, the total unfairness budget must be split between imbalance on the positive and negative classes. We build on \citet{pleiss2017fairness} by deriving the relationship between unfairness measures when calibration within groups is not constrained to hold perfectly and by characterizing how the total unfairness budget depends on accuracy in the context of a risk score for a binary classifier. \citet{bell2023possibility} study the impossibility theorem empirically, and find that allowing small violations can make approximate satisfaction of multiple fairness criteria feasible in practice, and that feasibility tends to expand with model performance. Our accounting identity illuminates these results theoretically by showing that higher predictive accuracy mechanically reduces the total amount of unfairness that must be distributed across competing forms of unfairness. More recently, \citet{zehlike2025beyond} develop tools from optimal transport to adjust algorithms in the direction of a user-specified prioritization of groupwise calibration and balance. Our approach is complementary: rather than proposing a particular adjustment procedure, we characterize the feasible combinations of unfairness types that any such procedure must satisfy.

Finally, a small but growing literature studies fairness in non-binary prediction settings (e.g., \citet{berk2017convex, agarwal2018reductions}). We contribute to this literature by providing an accounting identity that applies to natural extensions of fairness components from the binary setting. We also show why impossibility results derived for binary outcomes need not  extend to regression settings, while also identifying conditions under which similar trade-offs re-emerge.

The remainder of the paper proceeds as follows. Section \ref{sec:notation} describes our setting and defines our fairness terms. Section \ref{sec:binary} establishes and interprets the accounting identity, focusing on binary prediction tasks. Section \ref{sec:experiments} illustrates and explores the result on benchmark datasets. Section \ref{sec:nonbinary} focuses on non-binary prediction tasks and the applicability (or lack thereof) of the binary impossibility result. Section \ref{sec:classifiers} extends the accounting identity to classifiers rather than risk scores. Section \ref{sec:limitations} concludes by discussing limitations. Formal proofs for all results are provided in Appendix \ref{app:proofs}; details of experiments are in Appendix \ref{app:more_exp}.

\section{Problem Setting and Definitions}\label{sec:notation}
This section describes our problem setting and defines the fairness metrics on which we focus. 
\subsection{Problem Setting}
Individuals are characterized by a group, $G\in\{0,1\}$, an outcome, $Y\in\mathbb{R}$, and a vector of observable characteristics, $X\in\mathbb{R}^k$, that will be used to predict $Y$. Depending on the setting, $G$ may or may not be included in $X$. 

A \emph{risk score} $Z\in\mathbb{R}$ is a prediction for $Y$ based on the information in $X$. For example, when $Y$ is binary, $Z$ represents the predicted probability that $Y$ takes a value of $1$. In a regression setting, $Z$ represents the model's prediction for $Y$. We assume the existence of finite first and second moments for the joint distribution of $(Y,G,Z)$.

Our main substantive assumption is that $Z$ is globally calibrated:
 $$\E[Y|Z=z]=z \,\text{ for all } \,z.$$

Although our focus is on risk scores, we extend our results to classifiers in Section \ref{sec:classifiers}.

\subsection{Fairness Metrics} 

We focus on two types of unfairness with respect to the performance of an algorithm across groups.

\subsubsection*{Miscalibration Within Groups}~

Our first fairness metric is \emph{miscalibration within groups at a risk score $z$}, which we define as:
\begin{equation}\label{eq:delta_c_z_def}
   \delta_C(z) = \E[Y|Z=z,G=1]-\E[Y|Z=z,G=0].
\end{equation}
If a risk score is perfectly calibrated for each group, then the risk score is a sufficient statistic for the expected value of the outcome---information about group membership should provide no additional information, i.e., $(Y \perp G)|Z=z$. This implies $\E[Y|Z=z,G=1]=\E[Y|Z=z,G=0]$, so that $\delta_C(z)=0$. In contrast, $\delta_C(z)$ will be non-zero if, holding $Z$ fixed, individuals in one group tend to have higher values of $Y$ than individuals in the other group. The greater these differences are, the greater the magnitude of $\delta_C(z)$. 

By construction, $\delta_C(z)$ is a point-wise measure of an algorithm's miscalibration at a particular risk score $z$. It will also be convenient to define a summary measure of miscalibration across all risk scores:

\begin{equation}\label{eq:delta_c_def}
   \delta_C = \sum_z \, \delta_C(z) \, \omega_Z(z)
\end{equation}
where
\begin{equation}\label{eq:omega_z_def}
    \omega_Z(z) := \Var(G|Z=z)\,\text{P}(Z=z).
\end{equation}
In settings in which $Z$ is non-discrete, we define $\delta_C$ by substituting an integral into \eqref{eq:delta_c_def} in place of the summation and a density function for $Z$ in place of $\text{P}(Z)$ in \eqref{eq:omega_z_def}. In words, $\delta_C$ is a weighted sum of the within-score differences in the outcome across groups, where the weights reflect the extent to which both groups are represented at a given value of the risk score. In settings in which the direction of miscalibration varies across risk scores, $\delta_C$ summarizes the \emph{net} degree of miscalibration, accounting for differences in the group distribution at different risk scores.

A large literature treats calibration within groups as an important component of fairness (see \citet{corbett2023measure, hedden2021statistical,sep-algorithmic-fairness}). Often, the normative basis for this claim is that if the score is miscalibrated by group, then two people with the same predicted risk face different true risks depending on group membership, which in turn implies that decisions based on the algorithm are systematically less justified for one group than the other. In addition, when an algorithm is differentially miscalibrated by group, a decision-maker interested in the outcome being predicted will face an incentive to treat individuals differently based on group membership when that characteristic can be observed (i.e., to engage in statistical discrimination). 

Although miscalibration by group is usually discussed as an all-or-nothing property, algorithms that violate this property  may do so to differing degrees. In addition, the normative considerations that animate the concern with group-wise miscalibration tend to be larger the greater the degree to which an algorithm is miscalibrated by group. That is, if one is concerned with an algorithm not being perfectly calibrated by group, that concern would generally be larger the greater the degree to which the algorithm is miscalibrated. From this perspective, an appealing feature of $\delta_C(z)$, our measure of miscalibration within groups, is that it is non-binary, allowing it to reflect the continuous nature of the unfairness that occurs when an algorithm is miscalibrated within groups.

Finally, although miscalibration by group is generally discussed in the context of algorithms that predict a binary outcome,  the definition of $\delta_C(z)$ does not require $Y$ to be binary. As such, $\delta_C(z)$ provides a natural measure of miscalibration within groups in regression settings as well as in classification problems.

\subsubsection*{Imbalance Across Groups}~

Our second fairness metric captures differences by group in the risk scores of individuals with the same true outcome:
\begin{equation}\label{eq:delta_b_y_def}
    \delta_B(y) = \E[Z|Y=y,G=1] - \E[Z|Y=y,G=0].
\end{equation}
When risk scores are perfectly balanced across groups at some level of the outcome $y$, there is no systematic relationship among group membership and risk scores, i.e., $(Z \perp G) | Y=y$. In this case, $\E[Z|Y=y,G=1] = \E[Z|Y=y,G=0]$, so that $\delta_B(y)=0$. Whereas $\delta_C(z)$ measures the degree to which group membership predicts outcomes after conditioning on the risk score, $\delta_B(y)$ measures the degree to which group membership predicts risk scores after conditioning on the outcome. When $Y$ is binary, $\delta_B(0)$ measures the degree of imbalance for the negative class and $\delta_B(1)$ measures the degree of imbalance for the positive class.

As with calibration, it will be useful to define a summary measure of imbalance across levels of the outcome:
\begin{equation}\label{eq:delta_b_def}
   \delta_B = \sum_y \, \delta_B(y) \, \omega_Y(y)
\end{equation}
where
\begin{equation}\label{eq:omega_y_def}
    \omega_Y(y) := \Var(G|Y=y)\,\text{P}(Y=y).
\end{equation}
When $Y$ is continuous, we define $\delta_B$ by substituting an integral into \eqref{eq:delta_b_def} in place of the summation and a density function for $Y$ in place of $\text{P}(Y)$ in \eqref{eq:omega_y_def}. 
In words, $\delta_B$ is a weighted average of imbalance in the risk score across groups, where the average is taken over levels of the outcome, and where the weights reflect the frequency of outcome levels and the extent to which both groups are represented at the outcome level in question. For example, when $Y$ is binary, $\delta_B$ aggregates imbalance in both the positive class, $\delta_B(1)$, and the negative class, $\delta_B(0)$, weighting each by the share of the population in each class scaled by the prevalence of group overlap within that class. When $Y$ is non-binary, the interpretation of $\delta_B$ is the same, just averaged over more values of the outcome. In settings in which the direction of imbalance varies at different levels of the outcome, $\delta_B$ summarizes the \emph{net} degree of imbalance, accounting for differences in the group distributions at different levels of the outcome.

As with calibration, there is a large literature treating balance across groups as an important component of algorithmic fairness. And as with calibration, the normative considerations that support balance across groups are more implicated the greater is the degree to which an algorithm is imbalanced. Intuitively, balance across groups reflects a principle of horizontal equity: when a decision rule conditions treatment on the outcome being predicted, individuals who are alike in that relevant respect should be treated alike regardless of group membership. When one group systematically receives more favorable scores than another among individuals with the same outcome, decisions based on those scores impose unequal treatment that cannot be justified by differences in the target variable itself. The more that one group systematically receives more favorable risk scores than the other group after accounting for the outcome being predicted, the worse is this form of unfairness, and the larger in magnitude our measure of imbalance will be.

The notion of balance captured by $\delta_B(y)$ is closely related to the fairness criterion known as separation or equalized odds \citep{hardt2016equality,chouldechova2017fair,kleinberg2016inherent}. Equalized odds requires that a predictor be conditionally independent of group membership given the outcome—that is, $(Z \perp G)|Y=y$ for all $y$. When this condition holds exactly, we have $\delta_B(y) = 0$ for all $y$, and hence $\delta_B=0$ as well. In this sense, $\delta_B$ provides a continuous, scalar measure of the degree to which equalized odds is violated, aggregating conditional imbalances across outcome levels.

\section{The Accounting Identity}\label{sec:binary}

In this section we derive an equation linking miscalibration, imbalance, accuracy, and baseline differences in the distribution of the group by outcome. We initially derive results for the case in which $Y$ is binary. We return to the case of non-binary $Y$ in Section \ref{sec:nonbinary}.

Our first step is to establish a covariance representation for our summary measures of unfairness, $\delta_B$ and $\delta_C$. 

\begin{lemma}\label{lem:cov_delta_b_delta_c}
Define $\delta_C$ as in \eqref{eq:delta_c_def} and $\delta_B$ as in \eqref{eq:delta_b_def}. Then
   $\delta_B = \E[\Cov(Z,G|Y)]$
    \  and \ $\delta_C=\E[\Cov(Y,G|Z)]$
\end{lemma}

Lemma \ref{lem:cov_delta_b_delta_c} allows us to express our summary measures of point-wise miscalibration and point-wise imbalance as conditional covariances. The accounting identity then follows from successive applications of the law of total covariance: first decomposing $\Cov(Y,G)$ along the score $Z$, and then decomposing $\Cov(Z,G)$ along the outcome $Y$.

\begin{prop}\label{prop:accounting_binary}
    Suppose that $Y\in\{0,1\}$ and  $\,\E[Y|Z=z]=z$ for all $z$. Then
\begin{equation}\label{eq:accounting_bin}
  \underbrace{\delta_{B}(0)\,\omega_Y(0) + \delta_{B}(1)\,\omega_Y(1)}_{\delta_B} + \underbrace{\sum_z \, \delta_C(z)\,\omega_Z(z)}_{\delta_C}  = \text{MSE}(Z) \, \left(\text{P}(G=1|Y=1) - \text{P}(G=1|Y=0)  \right).
\end{equation}
\end{prop}

Proposition \ref{prop:accounting_binary} reveals a strikingly simple constraint that any globally calibrated risk score for binary prediction must satisfy. The right-hand side is a quantity we will refer to as the \emph{total unfairness budget}: it is large when (i) the score is inaccurate (high MSE) and (ii) group membership is strongly concentrated in one outcome class relative to the other. The left-hand side describes how this budget is allocated across three familiar forms of unfairness: imbalance on the negative class, $\delta_B(0)$, imbalance on the positive class $\delta_B(1)$, and miscalibration within groups, $\delta_C(z)$, each entering with a weight that reflects where the population mass and group overlap actually lie.

Several features of Proposition \ref{prop:accounting_binary} are worth highlighting. First, if we impose perfect calibration within groups and perfect balance for the positive and negative classes---$\delta_C(z)=0$ for all $z$ and $\delta_B(y)=0$ for $y\in\{0,1\}$---then the entire left-hand side of the constraint in Proposition \ref{prop:accounting_binary} is zero. The proposition then forces the right-hand side of the constraint to be zero as well, implying either $\text{MSE}(Z)=0$ (i.e., $Z$ is an oracle) or no baseline difference in outcome prevalence across groups (equivalently, equal base rates). \citet{kleinberg2016inherent}'s impossibility result thus follows from the Proposition as a direct corollary. More generally, Proposition \ref{prop:accounting_binary} establishes the geometry of the feasible fairness tradeoffs away from the benchmark case of perfect fairness that gives rise to the impossibility result. In particular, the Proposition shows how shifting one component of unfairness necessarily reallocates the others.

Second, the weights clarify \emph{where} unfairness matters for the identity. Imbalance at a particular outcome level contributes most when that outcome level is common and when both groups are substantially represented within it. Similarly, the weights emphasize miscalibration at risk scores where the two groups overlap. In sparse regions (few individuals) or nearly homogeneous regions (almost all one group), the corresponding unfairness term carries little weight and therefore uses little of the budget. Put differently, forms of unfairness in regions with little group overlap contribute less to the budget than forms of unfairness in regions where both groups are present.

Third, Proposition \ref{prop:accounting_binary} is a signed identity: unfairness in one direction can mechanically offset unfairness in another. This is not a normative claim that such offsets are ethically acceptable; it is a descriptive implication of the feasible set induced by global calibration. In applications where a decision-maker or regulator instead cares about (for example) the maximum imbalance at any outcome level, or the sum of absolute imbalances across levels, those criteria can be incorporated directly in an objective function---optimized subject to the constraint in Proposition \ref{prop:accounting_binary}.

Finally, Proposition \ref{prop:accounting_binary}  highlights a close link between accuracy and fairness. Rather than there being a generic tradeoff between these properties, the accounting identity highlights their complementary nature: holding fixed the problem environment, improving accuracy mechanically shrinks the total unfairness budget that must be allocated across the various forms of unfairness. Conversely, interventions that reduce an algorithm's accuracy may reduce certain forms of unfairness but only at the expense of increasing other forms of unfairness even more (at least in weighted net terms). We illustrate these dynamics in our experiments in the next section.

\section{Empirical Analysis}
\label{sec:experiments}

Here we provide a high-level overview of our empirical validation, including on the COMPAS dataset \cite{Compas_Dataset}; we provide additional details and results for Adult, German Credit, and Bank Marketing in Appendix \ref{app:more_exp}). We use dataset versions from AI Fairness 360 \cite{aif360} and exclude the protected attribute from training features. We use plug-in estimators for all quantities. Models are recalibrated via isotonic regression to ensure global calibration. We estimate $\hat{\delta}_C(z)$ and $\hat{\delta}_B(y)$ as sample covariances within bins (20 equal-frequency bins for $Z$). 

\textbf{Experiment 1: Finite-Sample Validity}
\label{sec:exp1} We verify that \eqref{eq:accounting_bin} holds approximately in finite samples using feature ablation. That is, we train models to predict the outcome with varying feature counts; as the number of features used increases, accuracy increases, reducing the total unfairness. We use logistic regression, decision trees, random forests, and gradient boosting.

\begin{figure}[!htb]
    \centering
\includegraphics[width=1.0\linewidth]{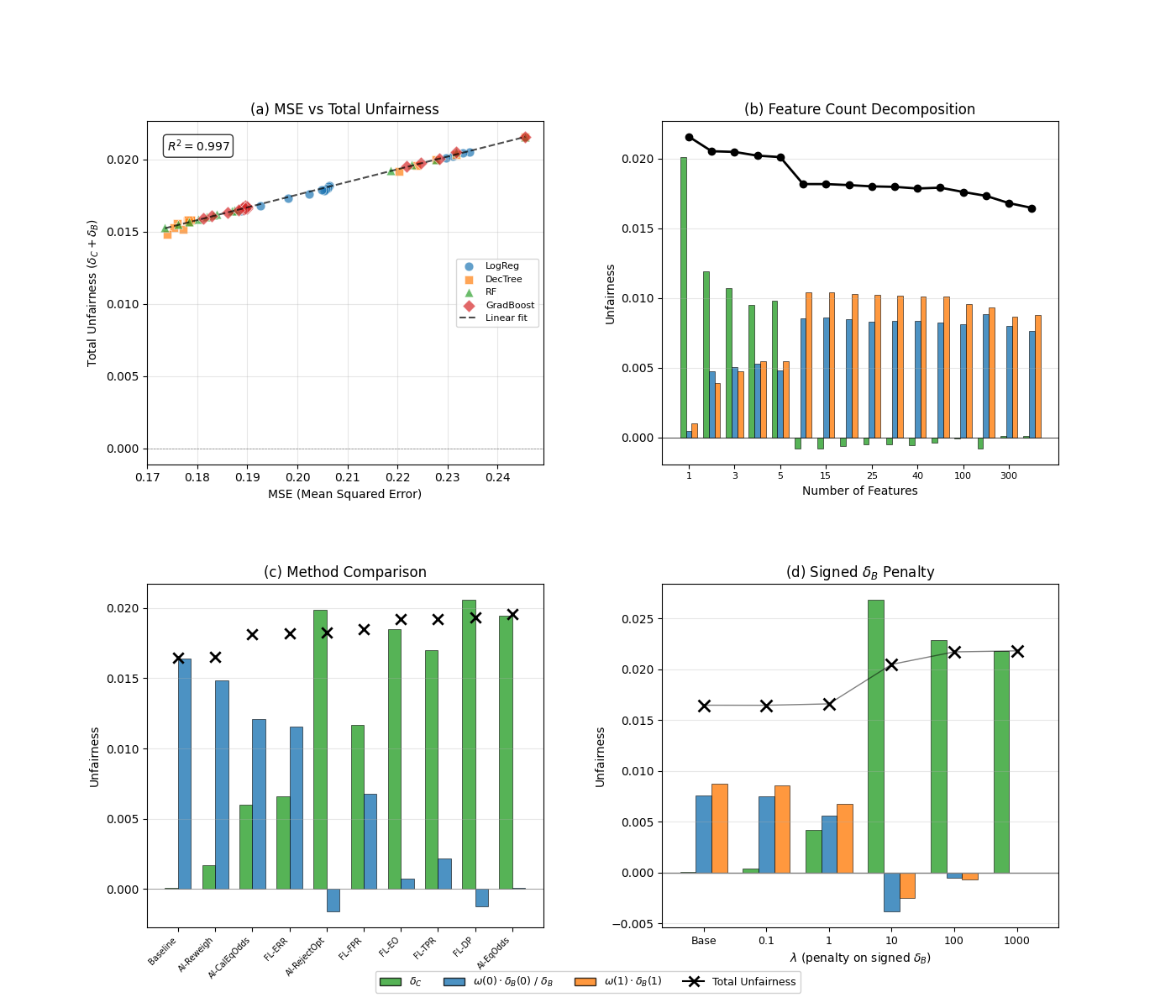}
    \caption{COMPAS experiments. \textbf{(a)} Identity validation: $\delta_B + \delta_C$ vs MSE for four classifiers with number of varying features included. \textbf{(b)} Decomposition of unfairness into constituent parts ($\delta_C$, $\delta_B(0)$ weighted by $\omega(0)$ and $\delta_C(0)$ scaled by $\omega(1)$ as estimated under each model) by feature count for logistic regression. \textbf{(c)} Decomposition of total unfairness among Fairness methods from AIF360/FairLearn. \textbf{(d)} Decomposition of unfairness from penalized regression varying $\lambda$ on $\delta_B$.}
    \label{fig:compas_all}
\end{figure}

Figure~\ref{fig:compas_all}(a) plots total unfairness $\hat{\delta}_B + \hat{\delta}_C$ against MSE. Empirical values closely track the theoretical prediction (dashed line). Panel (b) shows the three-way decomposition: as features increase, total unfairness (black line) decreases but the allocation across $\hat{\delta}_C$, $\hat{\delta}_B(0)$, and $\hat{\delta}_B$ shifts.

\textbf{Experiment 2: Fairness Methods Reallocate Unfairness}
\label{sec:exp3} We evaluate methods from FairLearn \cite{fairlearn} and AIF360 \cite{aif360}, plus penalized logistic regression targeting $\delta_B$. Figure~\ref{fig:compas_all}(c) shows that methods reducing one fairness metric tend to increase another while also increasing the total unfairness budget. Panel (d) varies the penalty $\lambda$ on $\delta_B(1)$: as $\lambda$ increases, imbalance decreases (as expected) but miscalibration grows, and total unfairness increases due to accuracy degradation.

\section{Prediction Tasks with Non-Binary Outcomes}\label{sec:nonbinary}

This section returns to the case in which the outcome $Y$ is non-binary (i.e., regression or non-binary classification). The following result provides a general form of the accounting identity.

\begin{prop}\label{prop:accounting_general}
    Suppose that $\E[Y|Z=z]=z$ for all $z$. Then
    \begin{equation*}\label{eq:accounting_gen}
     \underbrace{\sum_y\, \delta_B(y)\, \omega_Y(y)}_{\delta_B}  + \underbrace{\sum_z \, \delta_C(z)\,\omega_Z(z)}_{\delta_C} = \Cov(\E[G|Y] \,,\, Y-\E[Z|Y]).
    \end{equation*}
\end{prop}

Like Proposition \ref{prop:accounting_binary}, the accounting identity in Proposition \ref{prop:accounting_general} relates a weighted sum of each point-wise form of unfairness (the left-hand side of the constraint, equal to $\delta_B+\delta_C$) to a total unfairness budget (the right-hand side of the constraint). In the general case, the total unfairness budget is the covariance between the group prevalence at a given outcome level, $\E[G|Y]$, and the average prediction error at that outcome level, $Y-\E[Z|Y]$. 
Intuitively, it measures the association between where prediction error is concentrated along the outcome distribution and where group membership is concentrated along the outcome distribution. For example, the total unfairness against a disadvantaged group is larger precisely when that group is disproportionately concentrated at the levels of the outcome where the algorithm has systematically larger average prediction error. In contrast, total unfairness will be small when group prevalence does not systematically vary much by the outcome (so that $\E[G|Y]$ is nearly constant) or when the model's prediction errors tend to be similar across outcome levels (in which case $Z$ is close to being an oracle for $Y$).

Under the general form of the accounting identity, the key feature of an algorithm that determines the total unfairness budget is not simply the magnitude of its prediction errors---as it is in the binary outcome setting---but is rather how these prediction errors \emph{co-vary} with group membership across levels of the outcome. Consequently, the tight link between an algorithm's MSE and its total unfairness budget that we saw with binary outcomes need not extend to regression settings.

To see why, note that we can decompose MSE:
\begin{align}\label{eq:mse_decomp}
\E[(Y-Z)^2]
=\E\Big[\E\big[(Y-Z)^2\mid Y\big]\Big] 
=\Var\Big(Y-\E[Z\mid Y]\Big) \;+\; \E\Big[\Var(Z\mid Y)\Big].
\end{align}
The second term, $\E[\Var(Z\mid Y)]$, captures \emph{within-outcome noise} in the risk score among individuals with the same level of the outcome. An accuracy gain that comes purely from shrinking $\Var(Z\mid Y)$ will reduce MSE without affecting the total budget, $\Cov(\E[G|Y],\,Y-\E[Z\mid Y])$, since the total budget depends only on the systematic prediction errors at each outcome level. In contrast, an intervention that improves accuracy in a manner that reduces the spread of such errors across outcomes (i.e., that reduces the first term) can either increase or decrease the total fairness budget; the direction of the effect depends on whether it reduces prediction errors more at outcome levels with relatively high or relatively low group prevalence.

Thus, unlike in the binary setting, it is possible that increasing the accuracy of a regression algorithm could increase the total unfairness budget, and vice-versa. That being said, even in the non-binary setting, total unfairness remains quantitatively constrained by accuracy, as established in the following Proposition

\begin{prop}
\label{prop:accuracy_tightens_budget}
Suppose $Z$ is globally calibrated. Then: 
\begin{equation*}
\big|\delta_B+\delta_C\big|
\;\le\;
\sqrt{\Var(\E[G|Y])}\;\sqrt{\Var\big(Y-\E[Z\mid Y]\big)}\;\le\;
\sqrt{\Var(\E[G|Y])}\;\sqrt{MSE(Z)}.
\end{equation*}
\end{prop}

The loosening of the relationship between accuracy and fairness in the non-binary setting potentially relaxes the sharp impossibility results that characterize binary prediction problems, a possibility we explore in the next section. 

\subsection{Possibility Examples for Non-Binary $Y$}\label{sec:counterexample}
We now provide sufficient conditions under which a globally calibrated regressor can perfectly satisfy calibration within groups and error-rate balance pointwise, while being neither an oracle nor independent of group membership.

\begin{prop}[Sufficient conditions for pointwise fairness]
\label{prop:existence_suffic}
Let $W\subseteq X$ denote a subset of the available features. Define the score to be $Z=\E[Y|W]$. Suppose the following conditions are satisfied:
\begin{enumerate}
\item[(A1)] $\E[\Var(Y|W)]>0$.
\item[(A2)] $Y \perp G \mid W$.
\item[(A3)] $W \perp G \mid Y$.
\item[(A4)] $\Var(\E[G|W])>0$.
\end{enumerate}
Then: $Z$ is globally calibrated; $G \not\perp Y$; $Z$ is not an oracle; and for all $z$ and all $y$, $\delta_C(z)=\delta_B(y)=0$.
\end{prop}

In Proposition \ref{prop:existence_suffic}, global calibration follows from construction of $Z$. \emph{A1} guarantees that $Z$ is not an oracle. Conditions \emph{A2} and \emph{A3} restrict attention to settings in which the association between group membership and outcomes operates through the same channel as the association between group membership and the features used by the algorithm. In such settings, conditioning on either eliminates residual group information, guaranteeing group-wise calibration and balance. Finally, \emph{A1} provides that $G$ varies systematically with $Y$; in conjunction with \emph{A2} and \emph{A3}, this implies $G \not\perp Y$.

We now provide a stylized example to illustrate that the conditions in Proposition \ref{prop:existence_suffic} are not mutually incompatible. Let $Y\sim \text{Unif}[0,1]$. Let $X\in\{0,1\}$ be a coarse observable feature generated by $X:=\mathbf1\{Y>1/2\}$. Define the risk score $Z$ as the Bayes regressor: $Z=\E[Y|X]$, so that $Z=0.25$ if $X=0$ and $Z=0.75$ if $X=1$. By construction, $E[Y|Z=z]=z$ for each $z$, so global calibration holds. Because $Z$ takes only two values whereas $Y$ is non-binary, $Z$ is not an oracle. 

Assume that group membership depends only on the feature $X$: $\text{P}(G=1|X=0)=0.1$ and $\text{P}(G=1|X=1)=0.9$, with $(G \perp Y)|X$. Because $P(G=1|X)$ is increasing in $X$, and because $X$ is deterministically (weakly) increasing in $Y$, we know that $Y$ and $G$ are positively correlated.

To establish calibration within groups, note that $Z$ is a deterministic function of $X$. Hence, the fact that $(G \perp Y)|X$ implies $(G \perp Y)|Z$, which implies $\E[Y|Z,G]=E[Y|Z]$. Therefore, $\delta_C(z)=\E[Y|Z=z,G=1]-\E[Y|Z=z,G=0] = 0$.

To establish balance across groups, note that because $Z$ is a deterministic function of $X$, and $X$ is a deterministic function of $Y$, we have that $Z$ is a deterministic function of $Y$. Hence, $E[Z|Y,G]=E[Z|Y]$. Therefore $\delta_B(y)=\E[Z|Y=y,G=1]-\E[Z|Y=y,G=0] = 0$. 

Because groupwise calibration and balance respectively hold for each $z$ and $y$, the risk score perfectly satisfies pointwise calibration within groups and balance across groups. This is true despite the fact that $Z$ is not an oracle and there are baseline differences in $Y$ by $G$, establishing that the impossibility results do not directly extend to the non-binary setting.

\subsection{Sufficient Condition for Impossibility}
\label{sec:no-counterexample}

We now give a sufficient condition to restore the binary-style impossibility theorem for non-binary outcomes. The key obstacle in the regression setting is that, when $Y$ is non-binary, the right-hand side of Proposition \ref{prop:accounting_general} takes the form of a covariance between nontrivial functions of $Y$. Unlike in the binary case, this covariance can vanish through cancellation across different regions of the outcome distribution, even when the functions within the covariance are non-constant over $Y$. We show that a natural structural restriction---monotone outcome disparities combined with a reversion-to-the-mean property of prediction---rules out such cancellation and restores an impossibility-style result for a broad class of algorithms.

\begin{prop}
\label{prop:regression-impossibility}
Define $\pi(y)=\E[G|Y=y]$ and define $m(y)=\E[Z|Y=y]$. Suppose that: (1) $Z$ is globally calibrated; (2) $\pi(\cdot)$ is weakly monotonic (non-increasing or non-decreasing) on the support of $Y$; and (3) $m(\cdot)$ is 1-Lipschitz, i.e.: $$|m(y_2)-m(y_1)|\le |y_2-y_1|\quad \text{for all}\quad y_1,y_2.$$ Suppose the pointwise fairness measures $\delta_C(z)=0$ for all $z$ and $\delta_B(y)=0$ for all $y$. Then either $G\perp Y$, or $\text{MSE}(Z)=0$ (i.e., $Z$ is an oracle).
\end{prop}

Proposition \ref{prop:regression-impossibility} provides a regression-setting analogue to the binary impossibility result under additional structure. The first new assumption---that group prevalence varies monotonically across levels of the outcome---is determined by the problem setting and reflects the nature of systematic outcome disparities across groups. In contrast, the second new assumption---that the inverse regression function $m(y)=\E[Z|Y=y]$ is 1-Lipschitz---is a restriction on the behavior of the prediction rule itself. Intuitively, it requires predictions to exhibit a reversion-to-the-mean property, preventing $m(y)$ from amplifying variation in $Y$. Whether this condition holds in practice depends on the modeling choices, such as the degree of regularization imposed by the learning algorithm.

\section{Application to Classifiers}
\label{sec:classifiers}
This section extends our main results for risk scores to settings in which a binary classifier $\widehat{Y}\in\{0,1\}$ is taken as the primitive, along with binary $Y$ and $G$. The false positive and true positive rates by group are given by $FPR_g=\text{P}(\widehat{Y}=1|Y=0,G=g)$ and $TPR_g=\text{P}(\widehat{Y}=1|Y=1,G=g)$. The positive predicted value and false omission rates by group are $PPV_g=\text{P}(Y=1|\widehat{Y}=1,G=g)$ and $FOR_g=\text{P}(Y=1|\widehat{Y}=0,G=g)$.

Define $Z_{\widehat{Y}}=\text{P}(Y=1|\widehat{Y})$ to be the classifier-derived risk score. Because $\widehat{Y}$ is binary, $Z_{\widehat{Y}}$ takes only two values, which we refer to as $z_1=\text{P}(Y=1|\widehat{Y}=1)$ and $z_0=\text{P}(Y=1|\widehat{Y}=0)$. Hence $z_1$ corresponds to the classifier's (overall) PPV and $z_0$ to its (overall) FOR. By construction, $Z_{\widehat{Y}}$ is globally calibrated.

Because $Z_{\widehat{Y}}$ is a deterministic function of $\widehat{Y}$, conditioning on $Z_{\widehat{Y}}=z_1$ is equivalent to conditioning on $\widehat{Y}=1$, and similarly for $Z_{\widehat{Y}}=z_1$ and $z_0$. Hence, we can express our measure of within group calibration as
\begin{align}\label{eq:classifier_delta_c}
    \delta_C(z_1)=\E[Y|\widehat{Y}=1,G=1]-\E[Y|\widehat{Y}=1,G=0] = \Delta\,PPV. \\
    \delta_C(z_0)=\E[Y|\widehat{Y}=0,G=1]-\E[Y|\widehat{Y}=0,G=0] = \Delta \,FOR.
\end{align}
where $\Delta\,PPV = PPV_1-PPV_0$ and $\Delta\,FOR = FOR_1-FOR_0$. Using the definition of $Z_{\widehat{Y}}$, we can similarly express balance across groups as
\begin{align}
\delta_B(1)=\E[Z_{\widehat{Y}}|Y=1,G=1]-\E[Z_{\widehat{Y}}|Y=1,G=0] = (z_1-z_0)\,\Delta \,TPR. \\
\delta_B(0)=\E[Z_{\widehat{Y}}|Y=0,G=1]-\E[Z_{\widehat{Y}}|Y=0,G=0] = (z_1-z_0)\,\Delta \,FPR. \label{eq:classifier_delta_b}
\end{align}
where $\Delta\,TPR = TPR_1-TPR_0$ and $\Delta\,FPR = FPR_1-FPR_0$.

Applying Proposition \ref{prop:accounting_binary} along with these identities yields the following result.

\begin{prop}\label{prop:classifier}
Define $\omega_Y(y)=\Var(G|Y=y)\text{P}(Y=y)$ and $\omega_{\widehat{Y}}(\widehat{y})=\Var(G|\widehat{Y}=\widehat{y})\text{P}(\widehat{Y}=\widehat{y})$. Then for any binary classifier $\widehat{Y}$:
\begin{align*}MSE(Z_{\widehat{Y}})& \, (\text{P}(G=1|Y=1)-\text{P}(G=1|Y=0))  \\
   =& \underbrace{\omega_{\widehat{Y}}(1)\,\Delta\,PPV + \omega_{\widehat{Y}}(0)\,\Delta\,FOR}_{\delta_C} + \underbrace{(PPV-FOR)\,(\omega_Y(1)\,\Delta\,TPR + \omega_Y(0)\,\Delta\,FPR )}_{\delta_B}.  
\end{align*}
where $ MSE(Z_{\widehat{Y}}) =  \Var(Y|\widehat{Y})=\text{P}(\widehat{Y}=1,Y=0)\,PPV + \text{P}(\widehat{Y}=0,Y=1)\,(1-FOR)  $.
\end{prop}
The $PPV-FOR$ term in Proposition \ref{prop:classifier} reflects the classifier’s accuracy: how different the true class is between predicted positives and predicted negatives. If $\widehat{Y}$ is uninformative, $PPV\approx FOR$ and the balance across groups terms (i.e., the TPR and FPR gaps) don’t use up much of the total fairness budget. The proposition also highlights that the relevant measure of accuracy for determining a classifier's total unfairness budget is the MSE of the derived risk score, which can diverge from more common measures of a classifier's performance (e.g., overall accuracy or F1).

\section{Limitations}\label{sec:limitations}

Our analysis has several important limitations. First, our results are derived under the maintained assumption of global calibration, which is often imposed via post-processing and may not hold uniformly across subpopulations or under distribution shift. Second, there are many plausible fairness metrics beyond the ones we study; these would be governed by other constraints and may have a different relationship to accuracy. Relatedly, the fairness dimensions on which we focus can miss sources of unfairness arising from mis-measurement of features or labels or from interactions between the algorithm and other societal institutions that operate with bias. Third, we focus on a single binary protected attribute; extending the framework to multiple groups or intersectional attributes is an important direction for future work. Fourth, our accounting identity is descriptive rather than normative: it characterizes mechanical relationships among statistical fairness metrics, accuracy, and baseline disparities, but does not by itself determine which allocation of unfairness is ethically appropriate
in a given domain.

\newpage

\bibliographystyle{ACM-Reference-Format}
\bibliography{references}
\appendix
\section{Proofs}\label{app:proofs}
\begin{proof}
    \begin{proof}[Proof of Lemma \ref{lem:cov_delta_b_delta_c}]
Applying the definitions of $\delta_B$, $\delta_B(y)$, and $\omega_Y(y)$, we can write

\begin{align*}
\delta_B &= \sum_y \left(E[Z|Y=y,G=1]-E[Z|Y=y,G=0]\right) \Var(G|Y=y) P(Y=y) \\
&= \sum_y \Cov(Z,G|Y=y) P(Y=y) \\
&= \E[\Cov(Z,G|Y)].
\end{align*}
where the second equality follows from the fact that $G$ is binary and the third from the definition of expectation. The result for $\delta_C$ follows similarly. 
\end{proof}
\end{proof}
\begin{proof}[Proof of Proposition \ref{prop:accounting_general}]
    We first decompose $\Cov(Y,G)$ with respect to $Z$:
\begin{align}
\label{eq:cov_YA_Z}
\begin{split}
\Cov(Y,G) &= \E[\Cov(Y,G|Z)] + \Cov\left( \E[Y|Z],\,\E[G|Z] \right) \\
&= \delta_C + \Cov\left( Z,\,\E[G|Z] \right).
\end{split}
\end{align}

Next, we again apply the law of total covariance to decompose  $\Cov(Z,G)$ with respect to $Z$:
\begin{align}
\label{eq:cov_ZA_Z}
\begin{split}
\Cov(Z,G) &= \E[\Cov(Z,G|Z)] + \Cov\left( \E[Z|Z],\,\E[G|Z] \right) \\
&= 0 + \Cov\left( Z,\,\E[G|Z] \right).
\end{split}
\end{align}

Substituting \eqref{eq:cov_ZA_Z} into \eqref{eq:cov_YA_Z} yields
\begin{equation}\label{eq:cov_YA_2}
    \Cov(Y,G) = \delta_C + \Cov(Z,G).
\end{equation}

Next, we will decompose  $\Cov(Z,G)$ with respect to $Y$:
\begin{align}
\label{eq:cov_ZA_Y}
\begin{split}
\Cov(Z,G) &= \E[\Cov(Z,G|Y)] + \Cov\left( \E[Z|Y],\,\E[G|Y] \right).
\end{split}
\end{align}

Substituting \eqref{eq:cov_ZA_Y} into \eqref{eq:cov_YA_2} yields:
\begin{equation}\label{eq:cov_YA_3}
    \Cov(Y,G) = \delta_C + \delta_B + \Cov\left( \E[Z|Y],\,\E[G|Y] \right).
\end{equation}

A final application of the law of total covariance to $\Cov(Y,G)$ with respect to $Y$ yields
\begin{equation}\label{eq:cov_YG}
    \Cov(Y,G) = \Cov(Y,\E[G|Y]).
\end{equation}

Substituting \eqref{eq:cov_YG} into \eqref{eq:cov_YA_3} and rearranging yields:
\begin{equation}\label{eq:accounting_general_agg}
    \delta_B + \delta_C = \Cov(Y-\E[Z|Y] , \E[G|Y]).
\end{equation}
Finally, applying the definitions of $\delta_B$ and $\delta_C$  to \eqref{eq:accounting_general_agg} yields the general form of the accounting identity as reflected in the Proposition.
\end{proof}

\begin{lemma}\label{lem:varianceidentity}Suppose that $Y$ is binary and $Z=\E[Y|Z]$. Define $MSE \E[(Y-Z)]^2$. Then:
\begin{align*}
    MSE(Z)=\Var(Y)-\Var(Z).
\end{align*}
    
\end{lemma}
\begin{proof}
Expanding MSE yields:
\begin{align*}
    MSE(Z)=\E[Y^2-2YZ+Z^2]=\E[Y^2]-2\E[YZ]+\E[Z^2] =\E[\E[Y^2|Z]-2Z\E[Y|Z]+Z^2|Z].
\end{align*}
by the Law of Iterated Expectations. But since $Y$ is binary, $Y^2=Y$; by that and global calibration, we can write 
\begin{align*}
    \E\left[\E[Y^2|Z]-2Z\E[Y|Z]+Z^2\mid Z\right]=\E[\E[Y|Z]-2Z^2+Z^2]=\E[\E[Y|Z]-\E[Y|Z]^2]=\E[\Var(Y|Z)].
\end{align*}

By the Law of Total Variance and calibration:
\begin{align*}
    \Var(Y)=\E[\Var(Y|Z)]+\Var[\E(Y|Z)]=\E[\Var(Y|Z)]+\Var(Z);
\end{align*}
rearranging and substituting $MSE(Z)$ yields the desired equality.

\end{proof}

\begin{proof}[Proof of Proposition \ref{prop:accounting_binary}]Recall that the total unfairness budget in Proposition \ref{prop:accounting_general} is equal to $\Cov(Y-E[Z|Y] , E[G|Y])$. Because this term is the covariance of two functions of a binary random variable, we can write it as:
\begin{equation}\label{eq:cov_binary_1}
    \Cov(Y-\E[Z|Y] , \E[G|Y]) = \Var(Y)\, \left(\E[G|Y=1]-\E[G|Y=0]\right) \, \left(1 - (m_1-m_0)\right) 
\end{equation}
where $m_y=\E[Z|Y=y]$. With binary $Y$, we can also write 
\begin{equation*}
m_1-m_0 = \frac{\Cov(Y,Z)}{\Var(Y)} = \frac{\Var(Z)}{\Var(Y)}.
\end{equation*}
where the second equality follows from global calibration and an application of the law of total covariance to $\Cov(Y,Z)$ with respect to $Z$.

Define the mean-squared error for $Z$ as  $MSE(Z) = E[(Y-Z)^2]$. 

Using Lemma \ref{lem:varianceidentity}, we can write:
\begin{align}\label{eq:mse_y_z}
    MSE(Z)=\Var(Y)-\Var(Z).
\end{align}

Dividing both sides of \eqref{eq:mse_y_z} by $\Var(Y)$ yields: 

\begin{equation}\label{eq:mse_2}
\frac{MSE(Z)}{\Var(Y)}=1-\frac{\Var(Z)}{\Var(Y)} = 1-(m_1-m_0).
\end{equation}

Substituting \eqref{eq:cov_binary_1} and \eqref{eq:mse_2} into \eqref{eq:accounting_general_agg} yields:

\begin{equation}\label{eq:accounting_binary_agg}
 \delta_B + \delta_C = \text{MSE}(Z) \, \left(\E[G|Y=1] - \E[G|Y=0] \right).
\end{equation}

Applying the definitions of $\delta_B$ and $\delta_C$ for the case of $Y=1$ to \eqref{eq:accounting_binary_agg} yields the accounting identity for the case of binary $Y$.

\end{proof}

\begin{proof}[Proof of Proposition \ref{prop:accuracy_tightens_budget}]
The first inequality follows from Proposition \ref{prop:accounting_general} and Cauchy-Schwartz. The second follows from \eqref{eq:mse_decomp}.
\end{proof}

\begin{proof}[Proof of Proposition \ref{prop:existence_suffic}]
 
 We prove each claim in turn.
 \paragraph{(i) Global Calibration.}

By the law of iterated expectations:

$$\E[Y\mid Z]=\E\big(\E[Y\mid W]\mid Z\big)=\E[Z\mid Z]=Z,$$

so $Z$ is globally calibrated.

\paragraph{(ii) Pointwise calibration within groups: $\delta_C(z)=0$ for all $z$.}
It suffices to show that $\E[Y\mid Z,G]=Z$. Because $Z$ is a function of $W$, we can use iterated expectations to write
\begin{align*}
\E[Y\mid Z,G]
&=\E\big(\E[Y\mid W,Z,G]\mid Z,G\big) \\
&=\E\big(\E[Y\mid W,G]\mid Z,G\big).
\end{align*}
By (A2), $\E[Y\mid W,G]=\E[Y\mid W]=Z$. Hence
\[
\E[Y\mid Z,G]
=\E[Z\mid Z,G]
=Z.
\]
Therefore, for every $z$ in the support of $Z$ and each $g\in\{0,1\}$,
\[
\E[Y\mid Z=z,G=g]=z,
\]
and thus
\[
\delta_C(z)
=\E[Y\mid Z=z,G=1]-\E[Y\mid Z=z,G=0]
=0
\quad\text{for all }z.
\]

\paragraph{(iii) Pointwise balance across groups: $\delta_B(y)=0$ for all $y$.}
Write $Z=f(W)$ where $f(w):=\E[Y\mid W=w]$. Then
\[
\E[Z\mid Y,G]=\E[f(W)\mid Y,G].
\]
By (A3),
\[
\E[f(W)\mid Y,G]=\E[f(W)\mid Y].
\]
Hence
\[
\E[Z\mid Y,G]=\E[Z\mid Y],
\]
which implies that for every $y$ in the support of $Y$,
\[
\E[Z\mid Y=y,G=1]=\E[Z\mid Y=y,G=0],
\]
and therefore
\[
\delta_B(y)
=\E[Z\mid Y=y,G=1]-\E[Z\mid Y=y,G=0]
=0
\quad\text{for all }y.
\]

\paragraph{(iv) $Z$ is not an oracle.}
By the law of total variance and global calibration,
\[
\E[(Y-Z)^2]
=\E\big[(Y-\E[Y\mid W])^2\big]
=\E[\Var(Y\mid W)].
\]
Assumption (A1) states $\E[\Var(Y\mid W)]>0$, hence $\E[(Y-Z)^2]>0$, which implies $P(Z=Y)<1$. Thus $Z$ is not an oracle.

\paragraph{(v) $G\not\perp Y$.}
First note that (A2) implies $\E[G\mid W,Y]=\E[G\mid W]$ (conditioning on $W$, $Y$ carries no additional information about $G$).
Similarly, (A3) implies $\E[G\mid W,Y]=\E[G\mid Y]$ (conditioning on $Y$, $W$ carries no additional information about $G$).
Combining these yields
\[
\E[G\mid W]=\E[G\mid Y].
\]
Taking variances of both sides gives
\[
\Var(\E[G\mid Y])=\Var(\E[G\mid W]).
\]
By (A4), $\Var(\E[G\mid W])>0$, hence $\Var(\E[G\mid Y])>0$, which means $\E[G\mid Y]$ is not constant. Therefore $G$ is not independent of $Y$ (since $G\perp Y$ would imply $\E[G\mid Y]=\E[G]$).
\end{proof}

\begin{lemma}[Monotone functions have signed covariance]
\label{lem:monotone-cov-new}
Let $Y$ be nondegenerate and let $f(Y)$ and $g(Y)$ be square-integrable functions of $Y$.
If $f$ and $g$ are monotone in the same direction, then $\Cov(f(Y),g(Y))\ge 0$, with strict inequality if both are nonconstant.
If $f$ and $g$ are monotone in opposite directions, then $\Cov(f(Y),g(Y))\le 0$, with strict inequality if both are nonconstant.
\end{lemma}

\begin{proof}
Let $Y'$ be an independent copy of $Y$. Then
\[
\Cov(f(Y),g(Y))=\tfrac12\,\E\Big[(f(Y)-f(Y'))(g(Y)-g(Y'))\Big].
\]
If $f$ and $g$ are monotone in the same direction, the product is almost surely nonnegative; if in opposite directions, the product is almost surely
nonpositive. Strictness follows from nondegeneracy and nonconstancy.
\end{proof}

\begin{proof}[Proof of Proposition \ref{prop:regression-impossibility}]
Pointwise fairness implies the left-hand side of Equation~(14) is zero, hence by the accounting identity:
\begin{align} \label{eqn:prop3cond} 0 = \Cov\!\Big(\E[G|Y],\, Y-\E[Z\mid Y]\Big)=\Cov\!\Big(\pi(Y),\, r(Y)\Big),
\end{align}

where $\pi(Y):=\E[G\mid Y]$ and $r(Y):=Y-\E[Z\mid Y]$.

By assumption, $m$ is 1-Lipschitz; then $r(y)=y-m(y)$ must be monotone (nondecreasing) in $y$. To see this, suppose $y_2 \geq y_1$ and apply the Lipschitz definition:
\begin{align*}
    r(y_2)-r(y_1)&=y_2-m(y_2)-(y_1-m(y_1)) \\&= (y_2-y_1)-(m(y_2)-m(y_1)) \geq y_2-y_1-|y_2-y_1|=0. 
\end{align*}

In addition, $\pi(y)$ is monotone by assumption. Hence, by Lemma~\ref{lem:monotone-cov-new}, $\Cov(\pi(Y),r(Y))$ must have a fixed sign and is strictly nonzero if both functions are nonconstant.
Consequently, \eqref{eqn:prop3cond} implies that either $\pi(Y)$ is constant almost surely (hence $G\perp Y$) or $r(Y)$ is constant almost surely.

If $r(Y)$ is constant almost surely, then $\E[r(Y)]=0$ (since global calibration implies $\E[Z]=\E[\E[Y\mid Z]]=\E[Y]$), so the constant must be $0$,
and hence $r(Y)=0$ a.s. Therefore
\begin{equation}
\E[Z\mid Y]=Y \quad \text{almost surely}.
\label{eq:EZgivenY}
\end{equation}
Together with global calibration $\E[Y\mid Z]=Z$, we now show that $Z=Y$ almost surely. First,
\[
\E[YZ] =\E[\E[YZ|Z]]= \E\!\big[ Z\,\E[Y\mid Z]\big] = \E[Z^2],
\qquad\text{and}\qquad
\E[YZ] = \E\!\big[ Y\,\E[Z\mid Y]\big] = \E[Y^2],
\]
where the first equality uses $\E[Y\mid Z]=Z$ and the second uses \eqref{eq:EZgivenY}. Hence $\E[Y^2]=\E[Z^2]=\E[YZ]$, and thus
\[
\E[(Y-Z)^2] = \E[Y^2]-2\E[YZ]+\E[Z^2]=0.
\]
Therefore $Y=Z$ almost surely, i.e.\ $\Var(Y\mid Z)=0$, i.e., $MSE(Z)=0$.
\end{proof}

\begin{proof}[Proof of Proposition \ref{prop:classifier}]

The main result follows directly from applying equations \eqref{eq:classifier_delta_c}-\eqref{eq:classifier_delta_b} to Proposition \ref{prop:accounting_binary}. The expression for $MSE(Z_{\widehat{Y}})$ can be derived as follows. 

We first show that the mean squared error of $Z_{\widehat{Y}}$ equals the expected
conditional variance of $Y$ given $\widehat{Y}$:
\begin{align*}
\mathrm{MSE}(Z_{\widehat{Y}})
&:= \E\big[(Y-Z_{\widehat{Y}})^2\big] \\
&= \E\Big[\E\big[(Y-Z_{\widehat{Y}})^2 \mid \widehat{Y}\big]\Big] \\
&= \E\Big[\E\big[(Y-\E[Y\mid \widehat{Y}])^2 \mid \widehat{Y}\big]\Big] \\
&= \E\big[\Var(Y\mid \widehat{Y})\big].
\end{align*}

Next, compute $\E[\Var(Y\mid \widehat{Y})]$ by conditioning on the two values of
$\widehat{Y}$. Since $Y$ is Bernoulli conditional on $\widehat{Y}$, we have
\[
\Var(Y\mid \widehat{Y}=1)=PPV(1-PPV),
\qquad
\Var(Y\mid \widehat{Y}=0)=FOR(1-FOR).
\]
Therefore,
\begin{align*}
\E[\Var(Y\mid \widehat{Y})]
&= \Pr(\widehat{Y}=1)\Var(Y\mid \widehat{Y}=1) + \Pr(\widehat{Y}=0)\Var(Y\mid \widehat{Y}=0)\\
&= \Pr(\widehat{Y}=1)\,PPV(1-PPV) + \Pr(\widehat{Y}=0)\,FOR(1-FOR).
\end{align*}
Finally, rewrite each term using the joint probabilities of false positives and false
negatives. Note that
\[
\Pr(\widehat{Y}=1,Y=0)=\Pr(\widehat{Y}=1)\Pr(Y=0\mid \widehat{Y}=1)=\Pr(\widehat{Y}=1)(1-PPV),
\]
and
\[
\Pr(\widehat{Y}=0,Y=1)=\Pr(\widehat{Y}=0)\Pr(Y=1\mid \widehat{Y}=0)=\Pr(\widehat{Y}=0)\,FOR.
\]
Substituting these identities yields
\begin{align*}
\E[\Var(Y\mid \widehat{Y})]
&= \Pr(\widehat{Y}=1)\,PPV(1-PPV) + \Pr(\widehat{Y}=0)\,FOR(1-FOR)\\
&= \Pr(\widehat{Y}=1,Y=0)\,PPV + \Pr(\widehat{Y}=0,Y=1)\,(1-FOR).
\end{align*}
Combining with $\mathrm{MSE}(Z_{\widehat{Y}})=\E[\Var(Y\mid \widehat{Y})]$ proves the claim.
\end{proof}

\section{Detailed Overview of Experimental Results}\label{app:more_exp}

To validate our theoretical framework, we conduct two sets of experiments on four
well-studied fairness benchmark datasets. These experiments demonstrate that (1) the accounting identity holds in finite samples,
(2) improving accuracy reduces total unfairness rather than increasing it, and
(3) existing fairness interventions reallocate total unfairness across metrics rather than producing Pareto improvements. The datasets we use are COMPAS, Adult, German Credit, and Bank Marketing. For reproducibility, we use the versions of the dataset provided by the AI Fairness 360 Package \cite{aif360}, which includes standardized preprocessing steps (e.g. one-hot encoding categorical features and dropping rows with any null values) and the generation of protected/unprotected labels in accordance as commonly used throughout the literature. In all cases, we exclude the group variable of interest from features considered for model training.

\textbf{COMPAS.} The COMPAS dataset \cite{Compas_Dataset}, originally collected and released by ProPublica, provides data about defendants in Broward County, FL over the years 2013-2014. We designate the two year recidivism prediction as our outcome and race (with $G=1$ for African-American and $G=0$ for Caucasian) as our group variable. Additional features include defendant demographics like age, a description of the charge, and  number and severity of of prior counts, among others.

\textbf{Adult Income.} The Adult Income Dataset \cite{Adult_Dataset} contains individual level data from the U.S. census. We designate income being greater than \$50,000 as our outcome and sex (with $G=1$ for female and $G=0$ for male) as our group variable. Addtional features include age, education, occupation, type of work, among others. 

\textbf{German Credit.} The German Credit dataset \cite{German_Dataset} consists of records of individuals applying for loans. We designate credit risk status as our outcome and a binarized age (with $G=1$ for outside the 25-60 year range and $G=0$ otherwise) as our group variable. Additional features include age, job skill level, housing status, loan purpose, among others.

\textbf{Bank Marketing.} The Bank Marketing dataset \cite{Bank_Dataset} consists of records from a Portuguese bank around marketing a new account to customers. We designate opening an account as the outcome of interest ($Y$) and age ($G=1$ for above 25, $G=0$ otherwise) as our group variable. Additional features include education, marital status, existing loans, and job, among others.

\subsection{Estimating Sample Quantities}.
\label{sec:estimation}
To estimate sample analogues of our theoretical quantities, we use plug-in estimators.
For group-wise means of outcome $Y$, we define:
\begin{equation}
\hat{\mu}_g = \frac{\sum_{i: G_i = g} Y_i}{|\{i : G_i = g\}|}
\end{equation}

For a given model $f(X)$, we compute predicted probabilities $\hat{p}_i$ and apply isotonic regression to ensure global calibration. We then estimate the conditional covariances as:

$$\hat{\delta}_C(z) = \frac{1}{n_z} \sum_{i: \hat{p}_i \in B_z} y_i g_i - \frac{1}{n_z^2} \left(\sum_{i: \hat{p}_i \in B_z} y_i\right) \left(\sum_{i: \hat{p}_i \in B_z} g_i\right)$$

where $B_z$ denotes a bin of predicted probabilities centered at $z$, and $n_z$ is the number of observations in that bin. We use 20 equal-frequency bins for continuous predictions. The aggregate miscalibration is then $\hat{\delta}_C = \sum_z \frac{n_z}{n} \hat{\delta}_C(z)$.

Similarly, for imbalance:

$$\hat{\delta}_B(y) = \frac{1}{n_y} \sum_{i: y_i = y} \hat{p}_i g_i - \frac{1}{n_y^2} \left(\sum_{i: y_i = y} \hat{p}_i\right) \left(\sum_{i: y_i = y} g_i\right)$$

For binary outcomes, we compute this separately for $Y=0$ and $Y=1$ and aggregate as $\hat{\delta}_B = P(Y=1) \hat{\delta}_B(1) + P(Y=0) \hat{\delta}_B(0)$.

\subsection{Experiment 1: Finite-Sample  Validity via Feature Ablation}
\label{sec:exp1}

\textbf{Objective.} Our first experiment establishes that the accounting identity derived
in Equation~\eqref{eq:accounting_bin} holds approximately in finite samples typical of real-world
applications. We use \emph{feature ablation} to capture the effect of improving accuracy: we vary the number of features from just one to all, and train each model with access to those features. 

\textbf{Setup.} We train four standard classification algorithms on each dataset: logistic
regression, decision trees, random forests, and gradient boosting. For each model, we train a new instance for including each additional feature (where we treat each one-hot encoded variable as a new feature). For easy of deciphiring the figures, we include the models trained on the first few features and then multiples of 10 features. We incorporate features in the order they are listed in the dataset, except we exclude the sensitive feature. All models are recalibrated using isotonic regression to ensure global calibration holds.

\textbf{Results.} Panel (1) of each of Figures~\ref{fig:compas_all}-\ref{fig:german_all} plots, for each respective dataset, model, and number of features, the
estimated total unfairness $\hat{\delta}_B + \hat{\delta}_C$ against model mean squared error. The dashed black line shows the theoretical prediction from Equation~\eqref{eq:accounting_bin}.
Across all four datasets and all model types, the empirical values closely track the
theoretical prediction. In particular, as MSE decrease, total unfairness moves towareds 0 across each plot. 
Panel (2) in each of Figure~\ref{fig:compas_all}-\ref{fig:german_all} shows how the composition of unfairness changes as we increase the number of features. As the models are allowed to use increasing numbers of features, the
total unfairness decreases roughly monotonically, but the relative shares of $\delta_C$ and
$\delta_B$ shift, as well as sometimes their signs, shift.  
\subsection{Experiment 2: Fairness Methods Reallocate Unfairness}
\label{sec:exp3}

\textbf{Objective.} Our second experiment examines how existing fairness methods affect
the decomposition of unfairness. We evaluate methods from the popular \cite{fairlearn} and \cite{aif360} libraries. FairLearn began as an implementation of \cite{agarwal2018reductions}, and the library implements the methods described (among others) to produce fair classifiers using a reduction to particular cost-sensitive classification problems. In particular, we use the \emph{exponentiated gradient} algorithm to improve fairness metrics. AI360 implements other approaches, including preprocessing and post-processing. We also implement a direct approach by incorporating a penalty on $\delta_B$ in penalized logistic regression. 

\textbf{Setup.} For FairLearn, we focus on in-processing methods and use exponentiated gradient to fit fair classifiers with respect to demographic parity (FL-DP), equalized odds (FL-EO), true positive rate (FL-TPR), false positive rate (FL-FPR), and error rate parity (FL-ERR). For AIF360, we use the pre-processing reweighting approach (AI-Reweighing) and post-processing approaches that changes probabilities (AI-EqOdds, AI-CalEqOdds) or provide favorable treatment to minority observations with high uncertainty (AI-reject opt).

We also implement a penalized logistic regression
that directly targets $\delta_B$:
\begin{equation}
\arg\max_\beta \sum_i \left[ y_i \ln(\hat{p}_i) + (1 - y_i) \ln(1 - \hat{p}_i) \right]
- \lambda \hat{\delta}_B(\beta),
\end{equation}
and additionally 

\begin{equation}
\arg\max_\beta \sum_i \left[ y_i \ln(\hat{p}_i) + (1 - y_i) \ln(1 - \hat{p}_i) \right]
- \lambda \hat{\delta}_B(1,\beta),
\end{equation}
optimized using the SciPy \citet{scipy} library with a quasi-newton (L-GBFS-B) method. 

\begin{figure}
    \centering
    \includegraphics[width=1\linewidth]{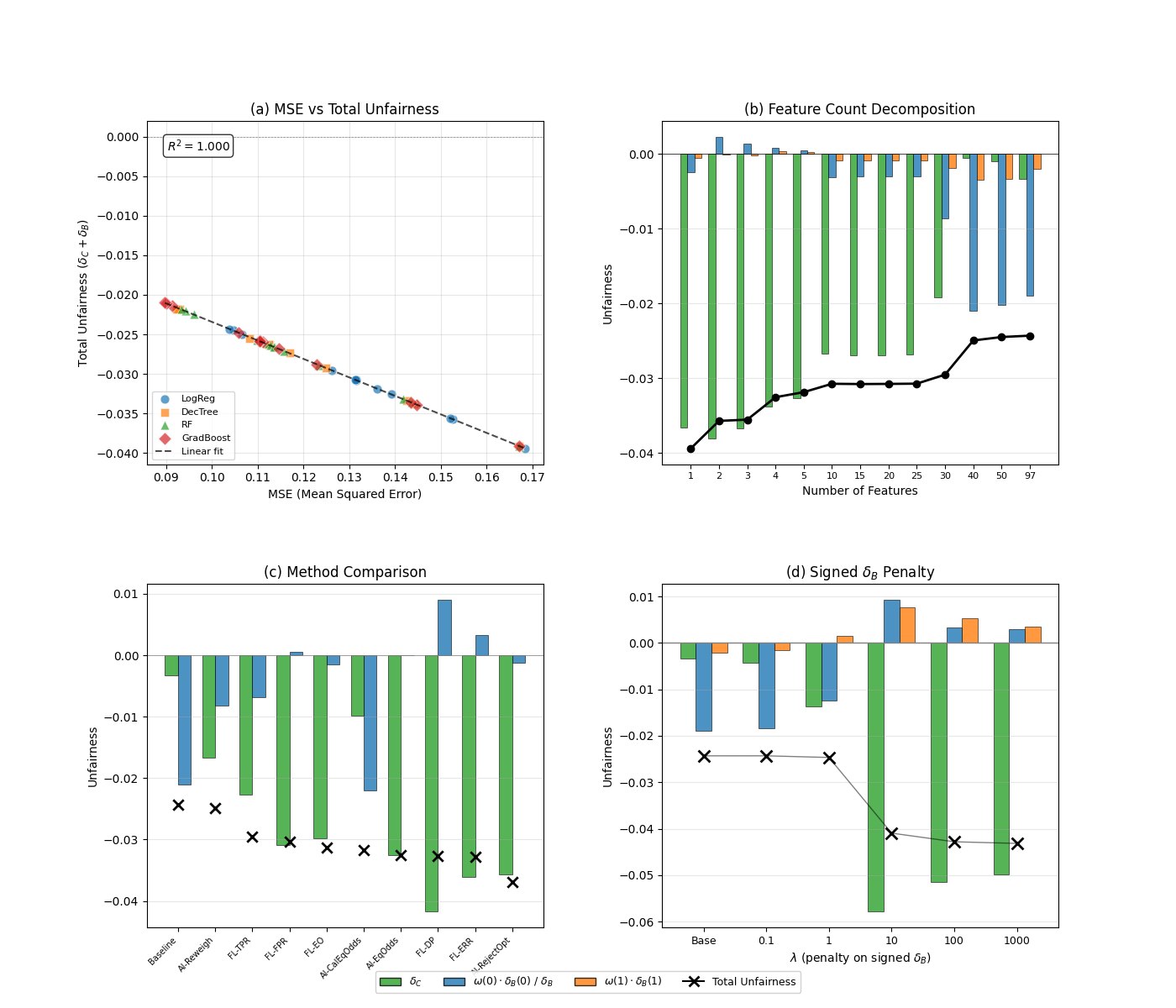}
    \caption{Adult experiments. \textbf{(a)} Identity validation: $\delta_B + \delta_C$ vs MSE for four classifiers with varying features. \textbf{(b)} Decomposition by feature count for logistic regression. \textbf{(c)} Fairness methods from AIF360/FairLearn. \textbf{(d)} Penalized regression varying $\lambda$ on $\delta_B(1)$.}
    \label{fig:adult_all}
\end{figure}

\begin{figure}
    \centering
    \includegraphics[width=1\linewidth]{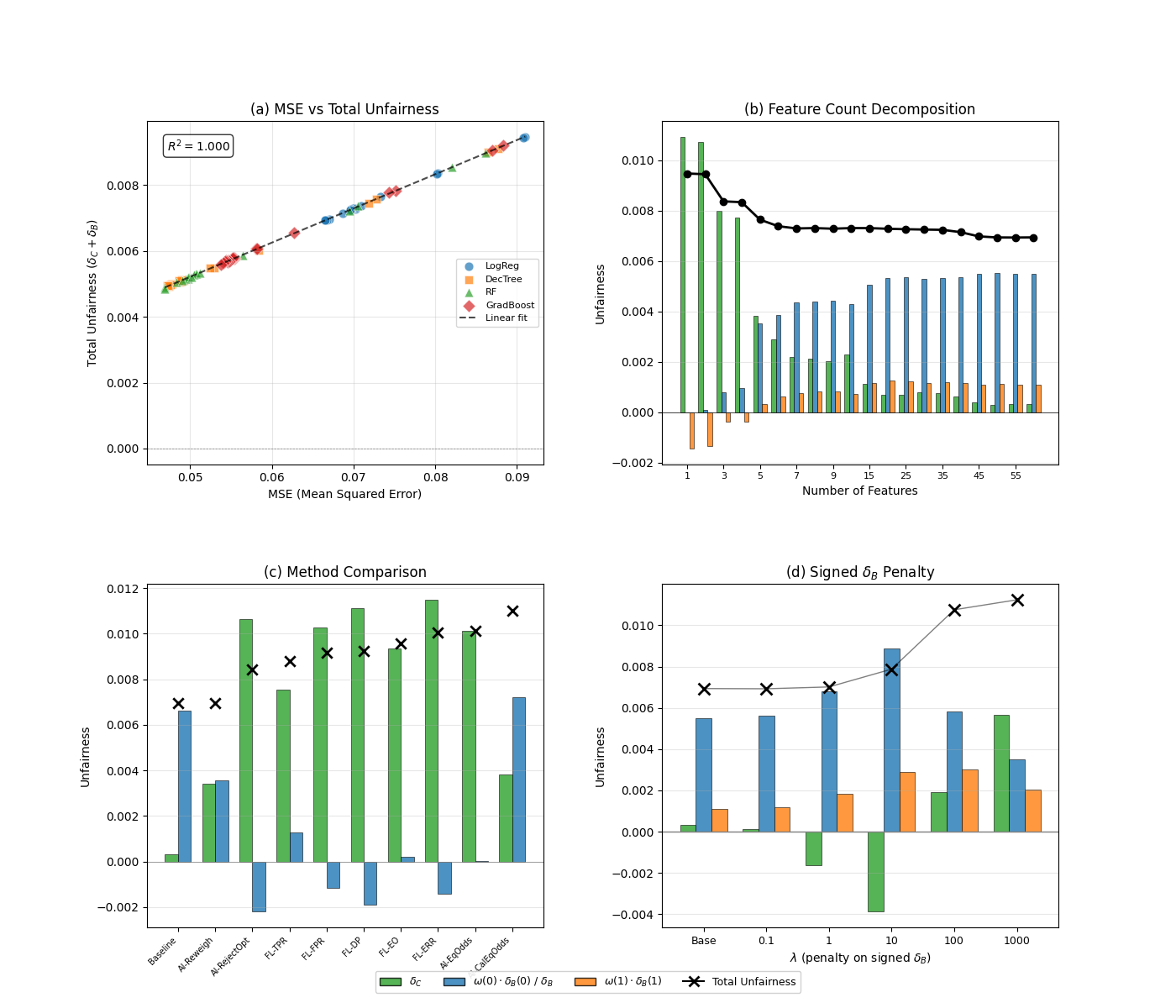}
    \caption{Bank experiments. \textbf{(a)} Identity validation: $\delta_B + \delta_C$ vs MSE for four classifiers with varying features. \textbf{(b)} Decomposition by feature count for logistic regression. \textbf{(c)} Fairness methods from AIF360/FairLearn. \textbf{(d)} Penalized regression varying $\lambda$ on $\delta_B(1)$.}
    \label{fig:bank_all}
\end{figure}

\begin{figure}
    \centering
    \includegraphics[width=1\linewidth]{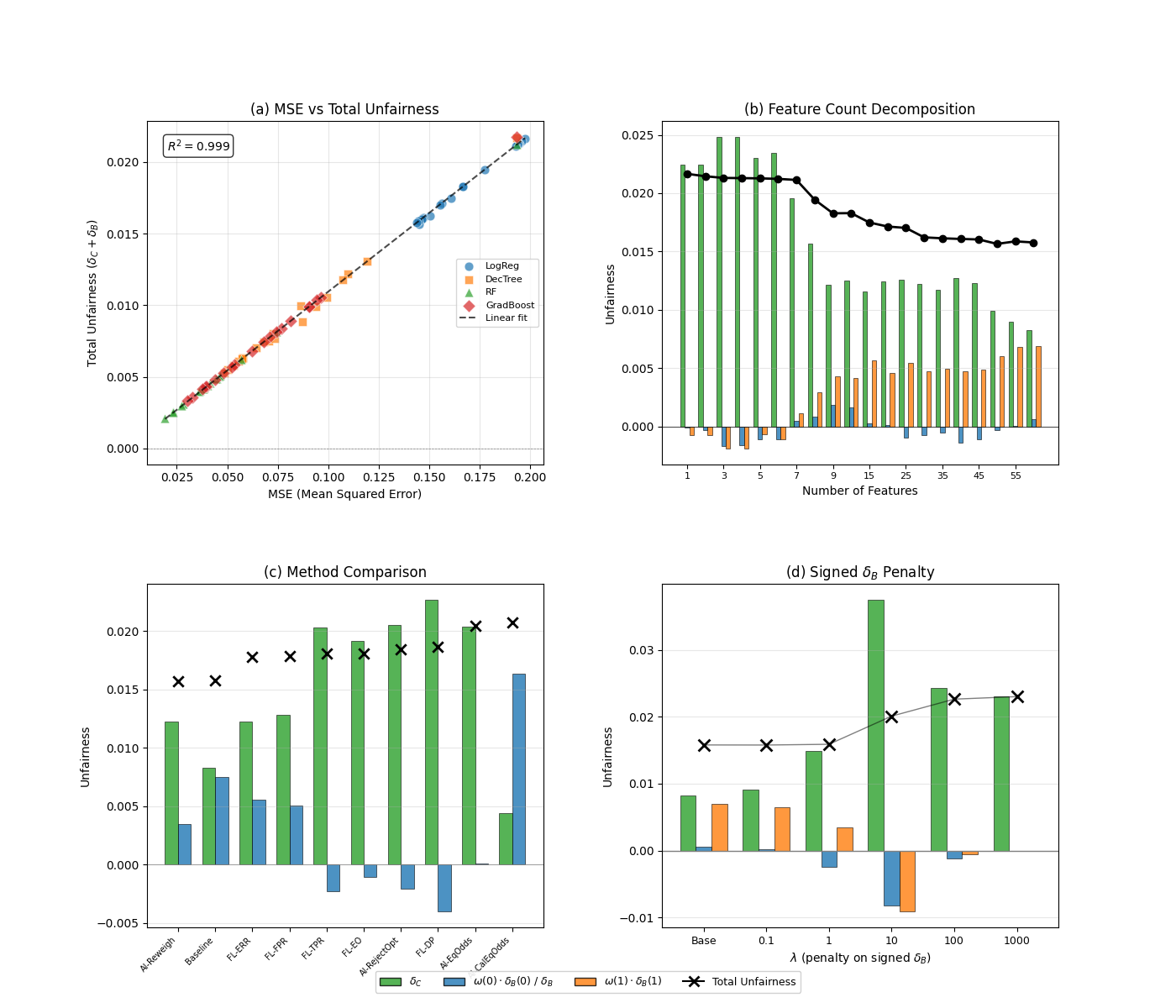}
    \caption{German experiments. \textbf{(a)} Identity validation: $\delta_B + \delta_C$ vs MSE for four classifiers with varying features. \textbf{(b)} Decomposition by feature count for logistic regression. \textbf{(c)} Fairness methods from AIF360/FairLearn. \textbf{(d)} Penalized regression varying $\lambda$ on $\delta_B(1)$.}
    \label{fig:german_all}
\end{figure}

\begin{figure}[!htb]
    \centering
    \includegraphics[width=1\linewidth]{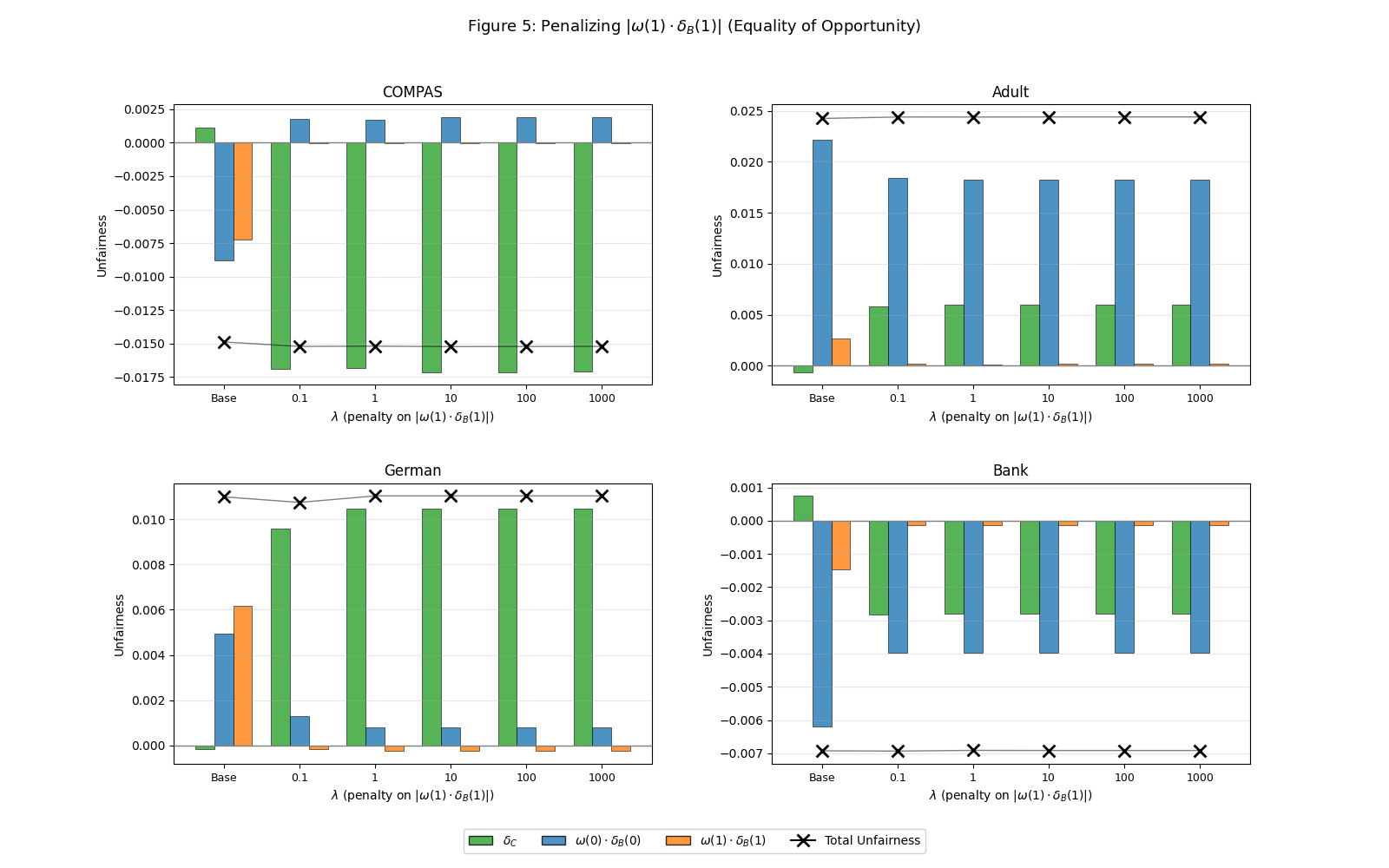}
    \caption{Unfairness decomposition for penalized regression approach, where the penalty is on $\delta_B(1)$ only. The height of each bar is the unfairness, split into $\delta_C$, $\delta_B(0)\omega(0)$, and $\delta_B(1)\omega(1)$, and the location of each bar corresponds to a different choice of lambda. The cross marks are the total unfairness predicted by equation \ref{eq:accounting_bin}.}
    \label{fig:delta_b1_penalty}
\end{figure}

\textbf{Results.} Panel (3) of Figures~\ref{fig:compas_all}-\ref{fig:german_all} compares the unfairness decomposition results
across fairness methods methods. Methods that successfully reduce one fairness metric tend to increase the
other. They also tend to reduce accuracy, which increases the total unfairness budget. Panel (4) of each figure focuses on the direct penalized logistic regression approach, and shows how the decomposition evolves as we increase
the penalty $\lambda$ on $\delta_B$. As $\lambda$ increases, imbalance decreases but
miscalibration grows, and the total budget increases because accuracy degrades. Figure \ref{fig:delta_b1_penalty} is analogous to panel (4) of each figure, but with the penalty only on $\delta_B(1)$ instead, corresponding to focusing on equality of opportunity. Notably, in some datasets, improving $\delta_B(1)$ seems to improve $\delta_B(0)$ in tandem, but for others (in particular, for COMPAS) improving $\delta_B(1)$ increases $\delta_B(0)$ to the point of flipping its sign.

\end{document}